\documentclass{article}

\pdfoutput=1

\usepackage{arxiv}

\usepackage[utf8]{inputenc} 
\usepackage[T1]{fontenc}    
\usepackage{url}            
\usepackage{booktabs}       
\usepackage{amsfonts}       
\usepackage{nicefrac}       
\usepackage{microtype}      
\usepackage{graphicx}
\usepackage[numbers]{natbib}
\usepackage{doi}

\usepackage{xcolor}
\usepackage{algorithm}
\usepackage{algorithmic}
\usepackage{amsmath}
\usepackage{multirow}
\usepackage{amsthm}

\title{Safety through Permissibility: Shield Construction for Fast and Safe Reinforcement Learning}

\author{ Alexander Politowicz \\
	Department of Computer Science\\
	University of Illinois Chicago\\
	Chicago, IL 60607 \\
	\texttt{politow2@uic.edu} \\
	\And
	  Sahisnu Mazumder\thanks{Work done while author was at the University of Illinois at Chicago.} \\
    Intelligent Systems Research \\
	Intel Labs\\
	Santa Clara, CA 95054 \\
	\texttt{sahisnumazumder@gmail.com} \\
    \And
    Bing Liu \\
	Department of Computer Science\\
	University of Illinois Chicago\\
	Chicago, IL 60607 \\
	\texttt{liub@uic.edu} \\
}

\date{}

\renewcommand{\shorttitle}{Permissibility and RL Safety}

\newtheorem{assumption}{Assumption}

\hypersetup{
pdftitle={Safety through Permissibility: Shield Construction for Fast and Safe Reinforcement Learning},
pdfsubject={rl, safety},
pdfauthor={Alexander Politowicz, Sahisnu Mazumder, Bing Liu},
pdfkeywords={RL Safety},
}

\begin{document}
\maketitle

\begin{abstract}
Designing Reinforcement Learning (RL) solutions for real-life problems remains a significant challenge. A major area of concern is safety. ``Shielding'' is a popular technique to enforce safety in RL by turning user-defined safety specifications into safe agent behavior. However, these methods either suffer from extreme learning delays, demand extensive human effort in designing models and safe domains in the problem, or require pre-computation. In this paper, we propose a new permissibility-based framework to deal with safety and shield construction. Permissibility was originally designed for eliminating (non-permissible) actions that will not lead to an optimal solution to improve RL training efficiency. This paper shows that safety can be naturally incorporated into this framework, i.e. extending permissibility to include safety, and thereby we can achieve both safety and improved efficiency. Experimental evaluation using three standard RL applications shows the effectiveness of the approach.
\end{abstract}


\section{Introduction}
\label{sec:intro}

Reinforcement learning (RL) \citep{sutton2018reinforcement} has made impressive progress recently on many difficult tasks including games \citep{Mnih2015} and robotics \citep{Wang2019}. However, it still requires significant improvements before it can be applied to real-life scenarios \citep{Dulac-Arnold2019}. Safety is one such area where the goal is to learn a policy that does not cause undesirable outcomes to either humans or the agent itself.
In standard RL, the agent learns only with the rewards it receives from the environment without any knowledge of what actions are safe. Thus, the learned optimal policy may involve unsafe actions that lead to unacceptable outcomes in safety-critical applications.
The topic of safe RL focuses on resolving this issue \citep{Garcia15}.

Shielding \citep{Alshiekh2018} is a standard technique in safe RL used to enforce safety by restricting any unsafe action selected by the agent. This constrained policy, known as a \emph{shield}, detects unsafe actions and re-selects safe actions instead. They are typically constructed using one of the three methods: first, rewards to the agent are altered to include a cost directly representing the safety of actions \citep{Garcia15}. The shield is implicitly learned as a part of the agent's policy. Second, a set of safe states is defined and a given or learned model is used to simulate trajectories from an action. Given such an action, if simulated trajectories take the agent out of this safe state set, the shield rejects the action as unsafe \citep{Bastani2019}. Third, safety specifications are provided to the agent by users, most often as linear temporal logic (LTL) properties \citep{Pnueli1977}, and a two-player safety game is solved to generate a set of safe states \citep{Alshiekh2018}. At each time step, the shield intercepts the action and rejects it if is not in this set. It is useful to note that shielding and learning in RL serve different functions. Shields are methods that prevent the agent from taking unsafe actions during learning, they do not guarantee that the agent will learn how to solve the problem. Interaction with the environment is still required for the agent to learn the optimal policy.

Current shield construction methods are limited. In the first method, the altered reward results in a different optimization objective. This changes the optimal policy and provides no guarantees for maintaining convergence in the original problem \citep{Garcia15}. The second method relies on models to construct the shield.
If the model is provided, users must spend additional effort designing and testing the model for correctness. If the model must be learned, additional complexity is added to the learning process resulting in slower convergence to the optimal policy \citep{moerland2020model,Garcia15}.
Furthermore, it is expected that the set of safe states is known and provided \citep{Giacobbe2021}. In the third method, pre-computation is required to convert the LTL properties into a shield. This involves a model of the problem dynamics, an automaton to process the LTL properties, and a method to run this automaton in the two-player safety game \citep{Alshiekh2018}. These steps require computation exponential in the state and action space dimensions, significantly limiting its usability \citep{Li2020}.

We propose a shield construction technique based on a framework called \emph{permissibility}, which was originally designed to improve RL training efficiency by cutting off entire regions of the exploration state space \citep{mazumder2018action}. An action is non-permissible if it can never lead to an optimal solution and thus should not be taken. Only permissible actions should be explored.
Safety can be naturally integrated into permissibility as unsafe actions also should not be taken.
That is, permissibility can be expanded to include safety as both unsafe actions and actions never leading to optimal solutions should not be taken in safe RL. By cutting off both sets, extended permissibility guarantees both safe behavior and improved training speed.

To connect safety and optimality in the framework of permissibility, \emph{we assume that users define the MDP such that unsafe transitions map to low rewards and episode termination}. 
Under extended permissibility, unsafe actions are a subset of non-permissible actions. None of these actions can be in either safe or optimal policies.
Permissibility-based shield construction diverges from previous shield construction methods because it directly combines agent learning of safe behaviors with learning of optimal policies. A shield is constructed by querying the set of unsafe and non-permissible actions defined by expanded permissibility. If the agent chooses an action in this set, the action is prevented from being executed.
Shields constructed using expanded permissibility neither change the optimization objective by altering rewards, nor require a model or knowledge of the MDP dynamics, and enable the guarantee of safe actions without any pre-computation. We argue that this method is easier for users to implement by allowing for more lenient design, facilitates greater generalization of safe agent behavior through learning, and increases optimal policy convergence speed by preventing actions leading to sub-optimal states, thus limiting necessary exploration.

Our contributions are twofold: 1) we show that safety can be efficiently implemented using an expanded permissibility definition, and 2) we demonstrate that shields implemented with expanded permissibility lead to improved RL learning efficiency.
Experiments are performed in three environments to illustrate these points.

\section{Related Work}
\label{sec:relwork}

Our work is inspired by \citet{mazumder2018action} and focuses on extending it to safe RL. Permissibility is related to shielding methods. Among recent shielding methods, those in~\citet{Alshiekh2018,Elsayed-Aly2021} construct shields through pre-computation of user-defined safety specifications. Those in \citet{Cheng2019,Jansen2020,Pranger2020,Bastani2019,Li2020,Giacobbe2021} use a given model to construct a shield by simulating agent actions and checking for safety. The work of \citet{junges2016safety} also focuses on safe RL but their ``permissibility'' is a different concept, referring to non-exclusivity in sets instead of safety and optimality.
All these prior works use safe state sets and either pre-computation or models to construct shields. By contrast, our permissibility-based shielding uses the definition of non-permissible actions to guarantee the learning of safe optimal policies. No pre-computation or model is needed, significantly reducing the amount of human work necessary for implementation.

Permissibility more broadly belongs to the realm of action elimination. \citet{Blanchi2004,Griffith2013,Seurin2020,Elimelech2020,Baram2021,Karimpanal2020} all learn to eliminate actions from heuristics, human feedback, or knowledge transfer.
\citet{Zahavy2018} propose Action Elimination Networks that learn and predict actions not to take.
\citet{Huang2020,Kanervisto2020} investigate how action elimination can benefit learning in different problem types. However, none of these works analyze action elimination with respect to safety while permissibility draws parallels between actions to be eliminated and unsafe actions.
Although permissibility does not fully automate the elimination of unsafe actions, it reduces the gap between user specification of safety and automatically learned safe behavior.

Among works that focus on limiting exploration in RL, \citet{Girgin2010,Bai2017} use trees and hierarchical models respectively to prevent sub-optimal actions. \citet{Zheng2006,Mahajany2017} use knowledge of the state space to either plan or limit exploration. \citet{Khetarpal2020} defines ``affordances'' in the context of RL and uses them to optimize exploration. \citet{Bourel2020} improve on existing upper confidence-based approaches. Permissibility is different as it not only limits exploration inherently but also ensures safety, which is the primary focus of this work.

\section{Background}
\label{sec:bkgd}

\paragraph{Reinforcement Learning} Given an interactable environment $\mathcal{E}$, we consider the episodic Markov Decision Process (MDP) setting of RL \citep{sutton2018reinforcement}, described by a tuple $M = (S, A, \tau, R, \gamma)$, where $S$ is the set of states, $A$ is the set of actions, $\tau$ is the transition function $\tau: S\ \times\ A \rightarrow S$, $R$ is the reward function $R: S\ \times\ A\ \rightarrow \mathbb{R}$, and $\gamma \in (0, 1]$ is the discount factor. Note that $A(s) \subseteq A$ denotes the set of actions available in state $s$. Both $S$ and $A$ can be either discrete or continuous. At any time step $t$ while in state $s_t$, the agent takes an action $a_t$, receives a scalar reward $r_t$, and reaches a new state $s_{t+1}$. This step of the RL agent composes a transition, given by the tuple ($s_t$, $a_t$, $r_t$, $s_{t+1}$). Tasks are limited to a certain number of time steps, constituting an ``episode''. Episodes terminate in either success (task completion) or failure (otherwise), subjectively defined by the user. A policy $\pi$ is a mapping from states to a probability distribution over actions $\pi: S \rightarrow \mathcal{P}(A)$ and can be either stochastic or deterministic. \emph{In our work, we consider only deterministic policies.} The return from a state is defined as the expected sum of discounted future rewards, $G_{t}=\mathbb{E}_{\pi}[\sum_{i=t}^T \gamma^{i-t} r_i]$, with $T$ as the horizon the agent optimizes over.
The goal of an RL agent is to learn an optimal policy $\pi^{*}$ that maximizes the expected return from the start distribution, i.e. $J = \mathbb{E}_{r_i, s_i\sim \mathcal{E}, a_{i}\sim \pi}[G_{t=0}]$.

\paragraph{Deep Q-learning}
Q-learning \citep{Watkins1992} employs the greedy policy $\pi(s)=\arg\max_{a} Q(s,a)$ to solve the RL problem. For continuous state spaces, Q-learning is performed with function approximators parameterized by $\theta^Q$, by minimizing the loss $L(\theta^Q)=\mathbb{E}_{s_t\sim \rho^\beta, a_t\sim \beta, r_i\sim \mathcal{E}}$ [($Q(s_t, a_t) - y_t)^2$], where $y_t=r(s_t, a_t) ~+ \gamma Q(s_{t+1},\pi(a_{t+1}))$ and $\rho^\beta$ is the discounted state transition distribution for some policy $\beta$. The dependency of $y_t$ on $\theta^Q$ is typically ignored. \citet{Mnih2015} adapted Q-learning by using deep neural networks as non-linear function approximators and a replay buffer to stabilize learning, known as Deep Q-Network or DQN. \citet{van2016deep} introduced Double Deep Q-Network (DDQN) by including a separate target network for calculating $y_t$ to deal with DQN's over-estimation problem.
For continuous action space problems, Q-learning is usually solved using an Actor-Critic method, e.g. DDPG \citep{Lillicrap2016}. DDPG maintains an Actor $\pi(s)$ with parameters $\theta^\pi$, a Critic $Q(s, a)$ with parameters $\theta^Q$, and a replay buffer that stores transitions for training. Training rollouts are collected with extra noise for exploration: $a_t = \pi(s)+\mathcal{N}_t$, where $\mathcal{N}_t$ is a noise process.

\paragraph{Shielding} Let $S_{\text{safe}} \subseteq S$ be a set of \emph{safe states}. As RL safety is application-dependent, the set of safe states is user-defined. Safe RL formulates safety as transitions $(s_t, a_t, r_t, s_{t+1})$ such that $s_{t+1} \in S_{\text{safe}}$.
The goal is to learn an optimal policy while ensuring that agents always stay within $S_{\text{safe}}$. Typically, this is achieved through a shield $\pi_{\text{safe}}$, detailed below
\begin{equation}
  \pi_{\text{safe}}(s_t) =
  \begin{cases}
                        \pi_{\text{shield}}(s_t) & \text{if}\ s_{t+1} \notin S_{\text{safe}} \\
                        \pi(s_t) & \text{otherwise} \\
  \end{cases}
\end{equation}
where $\pi$ is the agent's learned policy and $\pi_{\text{shield}}$ is a backup policy \citep{Bastani2019}.

Shields can be classified as either preemptive or post-posed \citep{Alshiekh2018}. In \textbf{preemptive shielding}, given a state $s_t$, the shield computes the set of safe actions $A_{\text{safe}}(s_t)$ which is then provided to the agent for action selection $a_t \sim \pi_{\text{safe}}(s_t) \in A_{\text{safe}}(s_t)$.
Note that $A_{\text{safe}}(s_t) = \{a \ | \ a \in A(s_t) \ \wedge\ \tau(s_t, a) \in S_{\text{safe}}\}$ is the set of \emph{safe actions}. In \textbf{post-posed shielding}, given an action $a_t \sim \pi(s_t)$ selected by the agent, the shield either allows $a_t$ if it is safe or re-selects a safe action $a_t \sim \pi_{\text{shield}}(s_t)$.

The problem of shielding is to construct the policy $\pi_{\text{safe}}$ and its components according to the problem under consideration.
Refer to Section \ref{sec:relwork} for details on existing shield construction methods.
A shield should satisfy two properties: 1) \emph{guaranteed correctness}, and 2) \emph{minimal interference}.
Guaranteed correctness requires that the shield ensures the agent satisfies the user-defined safety specification at all times.
Minimal interference requires that the shield use $\pi_{\text{shield}}$ as little as possible.

\paragraph{Permissibility} \citet{mazumder2018action} describe permissibility in the space of states and actions. Given a state $s_t$, an action $a_t$ is said to be \emph{non-permissible} in $s_t$ if it cannot lead to optimal expected return. If $a_t$ in $s_t$ is not known to be non-permissible, then it is \emph{permissible}.

There are two types of permissibility. \textbf{Type 1 permissibility} is when, given some transition $(s_t, a_t, r_t, s_{t+1})$, there is a function $f_1: S\ \times\ A\ \times\ S \rightarrow \{0, 1\}$. $f_1$ indicates if an action $a_t$ in state $s_t$ is \emph{permissible} ($f_1(s_t,a_t,s_{t+1})=1$) or \emph{non-permissible} ($f_1(s_t,a_t,s_{t+1})=0$) after transitioning to $s_{t+1}$. \textbf{Type 2 permissibility} is when, given an identically structured transition, there is a function $f_2: S\ \times\ A \rightarrow \{0, 1\}$. $f_2$ indicates if an action $a_t$ in state $s_t$ is \emph{permissible} ($f_2(s_t,a_t)=1$) or \emph{non-permissible} ($f_2(s_t,a_t)=0$) without performing $a_t$.
Type 1 permissibility defeats the purpose of safety because it requires the agent to have already transitioned to $s_{t+1}$ even if it may be unsafe. Furthermore, as the set of permissible states is provided, using Type 1 permissibility would be analogous to previous shielding methods that use provided sets of safe states.
\textbf{As such, we focus on Type 2 permissibility in this paper.}

\section{Permissibility-Based Shielding}
\label{sec:method}

Let $A_{\text{unsafe}}(s)$ be the set of unsafe actions in state $s$, given by $A_{\text{unsafe}}(s) = A(s) - A_{\text{safe}}(s)$, and $A_{p}(s)$ ($A_{np}(s)$) the set of permissible (non-permissible) actions in state $s$, defined under the Type 2 permissibility function $f_2$. We detail the permissibility-based shielding approach below.

\subsection{Unifying Safety and Permissibility}

Our goal is to connect unsafe actions with non-permissible actions. For our proposed method to work, we need to make the following assumption. 

\begin{assumption}
For an MDP $M = (S, A, \tau, R, \gamma)$, let $r_{\text{min}} = \min(R)$ be the minimum possible reward and $S_{\text{F}} \subset S$ be the set of final (terminal) states in $M$. Given that failure transitions receive $r = R_{\text{min}}$ and all other non-failure transitions receive $r > R_{\text{min}}$, we assume that $M$ treats unsafe actions as failure transitions, assigning them minimal rewards and terminating the episode when they occur
\begin{align*}
    \forall a \in A_{\text{unsafe}}(s),\ & R(s,a,\tau(s,a)) = r_{\text{min}}\ \wedge\\
    & \tau(s,a) \in S_F
\end{align*}
for some state $s$.
\label{assump}
\end{assumption}

Assumption \ref{assump} is reasonable because taking an action $a \in A_{\text{unsafe}}(s)$ is unsafe and naturally should receive minimal reward and result in failure episode termination. As both safety and permissibility indicate actions which should not be taken, it is intuitive to directly expand Type 2 permissibility to include safety. Assumption \ref{assump} connects unsafe and non-permissible actions because, by the Bellman update \citep{sutton2018reinforcement}, a transition resulting in $r_{\text{min}}$ and termination must have sub-optimal return. By imposing these conditions on $A_{\text{unsafe}}(s)$, we have satisfied the condition for non-permissibility.

Let $A_{unp}(s)$ be the set of unsafe non-permissible (``UNP'') actions and its complement, $A_{sp}(s)$, be the set of safe permissible (``SP'') actions for state $s$, as given by
\begin{equation}
\begin{aligned}
    & A_{unp}(s) = A_{\text{unsafe}}(s) \cup A_{np}(s)\\
    & A_{sp}(s) = A_{\text{safe}}(s) \cap A_p(s)
\end{aligned}
\end{equation}
We can use Assumption \ref{assump} to define $A_{unp}(s)$ and $A_{sp}(s)$. Then, we can treat $A_{unp}(s)$ as $A_{np}(s)$ and $A_{sp}(s)$ as $A_p(s)$ and directly apply the permissibility-based RL methods proposed in \citet{mazumder2018action}. This results in solving the RL safety problem while also achieving efficient training.

\subsection{Unsafe Non-Permissibility Shield Construction}

UNP shield construction requires user specification of $A_{unp}(s)$. When the agent is in state $s$ and has selected action $a$, the shield checks if $a \in A_{unp}(s)$. If so, the shield rejects the action and re-selects a safe one.

To simplify design of $A_{unp}(s)$, $s$ can be unambiguously defined using feature representations of the state space, 
where we represent a state $s \in S$ with an $n$-dimensional feature vector $\phi(s)=\{x_{1}, \dots, x_{n}\} \in \mathbb{R}^n$. Each feature $x_{i}$ can be either discrete or continuous.
The designer of the RL problem can analyze the value range of each feature dimension and define $A_{unp}(\phi(s))$. We discuss and provide examples of defining $A_{unp}(\phi(s))$ in the experiments.

\paragraph{Shield Construction} UNP shield construction relies only on $A_{unp}(s)$ for some $s$. Given a state $s$ and an action $a$, the shield is constructed as such:
\begin{equation}
  \pi_{unp,\text{safe}}(s) =
  \begin{cases}
                        \pi_{unp, \text{shield}}(s) & \text{if}\ \pi(s) \in A_{unp}(s) \\
                        \pi(s) & \text{otherwise} \\
  \end{cases}
\end{equation}
The next step in UNP shield construction is defining the backup policy $\pi_{unp, \text{shield}}$. As $A_{sp}(s) \subseteq A_{\text{safe}}(s)$, any policy that selects actions from $A_{sp}(s)$ must be safe. We show in the experiments some examples of backup policies.

\subsection{Discussion}
\label{sec:method_disc}

\paragraph{Shield Property Satisfaction} Shields constructed using UNP fall under post-posed shielding \citep{Alshiekh2018}. We show how well they satisfy the two properties required of shields, that is: \textbf{guaranteed correctness} and \textbf{minimal interference}. Note that $A_{\text{unsafe}}(s) \subseteq A_{unp}(s)$ given Assumption \ref{assump}. As the shields block any transitions involving $a \in A_{unp}(s)$, they also block all unsafe actions. By definition, UNP shields satisfy the property of guaranteed correctness. UNP shields also satisfy the property of minimal interference. Not only are unsafe actions blocked but actions generalized to be unsafe via Assumption \ref{assump} and RL \citep{sutton2018reinforcement} are also blocked, covering unsafe cases not specified by users. In fact, UNP shields are more powerful than previous shields because they also prevent sub-optimal actions $a \in A_{np}(s)$. All other safe and permissible actions are allowed.

\paragraph{Advantage of Action-Based Shielding} Note the distinction between the focus on states ($S_{\text{safe}}$) in previous shield construction works and actions ($A_{unp}(s)$) in our proposed method. Previous shielding methods give the agent $S_{\text{safe}}$ and require that the agent know if an action $a_t$ in a state $s_t$ will lead to $s_{t+1} \notin S_{\text{safe}}$. There is an implicit assumption that the shield knows $s_{t+1}$ \emph{before} $a_t$ is taken. This requires either knowledge of $\tau$ or access to an accurate model that can perform this prediction. In practice, this is often not feasible.
Shields that check if an action belongs to $A_{\text{unsafe}}(s)$ avoid these issues. In fact, we argue that action-based shielding more accurately matches intuitive notions of safety. In real-life, it is more intuitive for \emph{actions} to be classified as unsafe instead of states (i.e. $S_{\text{unsafe}} = S - S_{\text{safe}}$), which are perhaps more appropriately classified as ``dangerous''.
If agents happen to enter $S_{\text{unsafe}}$, action-based shields ensure that no unsafe actions occur thereafter. This fulfills safety guarantees at a greater resolution than previous state-based shielding methods. Furthermore, unsafe actions are eliminated without requiring the agent to know $s_{t+1} \in S_{\text{unsafe}}$.
In summary, action-based shields eliminate the state prediction problem in favor of a simpler action lookup. UNP allows for the construction of action-based shields by defining $A_{unp}(s)$.

\paragraph{Non-Comprehensive Safety Specifications} It is important to understand that our proposed method of shield construction is not meant to fully eliminate human work in the problem. RL safety requires user specification of fundamental safety information. Our claim is simply that previous methods of shield construction have required \emph{more} work than our idea of UNP. Identifying LTL properties, calculating safety invariants \citep{anderson20}, specifying a model, etc. all require significant amounts of domain knowledge and human work. A simplified comparison between our UNP shield construction method and previous shield construction methods is presented in Table \ref{table1}. An important note is that we \emph{do not} require $A_{unp}(s)$ be specified for all $s \in S$.
By combining unsafe actions and task failure through UNP, we have helped bridge the gap between generalization in RL and manual user safety specification, thus eliminating the need for comprehensiveness.
This shifts shield construction work to agent learning and significantly reduces the amount of human work needed. Agents can now take any UNP experience and generalize unsafe (or non-permissible) behavior across the state-action space.
We show in our experiments how specifying simple, non-comprehensive $A_{unp}(s)$ takes little time and still results in optimal \emph{and} safe behavior.

\begin{table}[ht]
\centering
\begin{tabular}{c|l}
    \multirow{3}{7em}{\emph{Pre-Computation}} 
    & Less required human work \\ 
    & Generalizes unsafe behaviors \\ 
    & Doesn't require $\tau$ knowledge or model \\ 
    \\
    \multirow{2}{7em}{\emph{Model-Based}} 
    & Less required human work \\ 
    & No waiting for model computations \\ 
    \\
    \multirow{2}{7em}{\emph{Safety Cost}} 
    & Doesn't change the solution \\ 
    & Easier to design \\ 
\end{tabular}
\caption{Advantages of UNP shield construction over previous shield construction techniques.}
\label{table1}
\end{table}

\paragraph{Limitations} We limit our analysis of UNP to deterministic MDPs. Deterministic environments encompass many realistic problems and have been solved using RL with great success \citep{Mnih2015,Wang2019}. Other types of RL problem formulations, such as partial observability, dynamic or non-stationary environments, etc., require significant additional design. 
UNP shield construction extends to the multi-dimensional action space and multi-agent domains with few adjustments. In the case of multi-dimensional action space, the user needs to analyze each dimension of the action space for UNP. For multi-agent problems, identification of $A_{unp}(s)$ must be done for each agent but the procedure remains similar. More creative ways of adapting permissibility is an interesting challenge and left for future work.

\section{Experiments}
\label{sec:exp}

\begin{figure*}[t]
  \centering
  \includegraphics[width=0.95\textwidth]{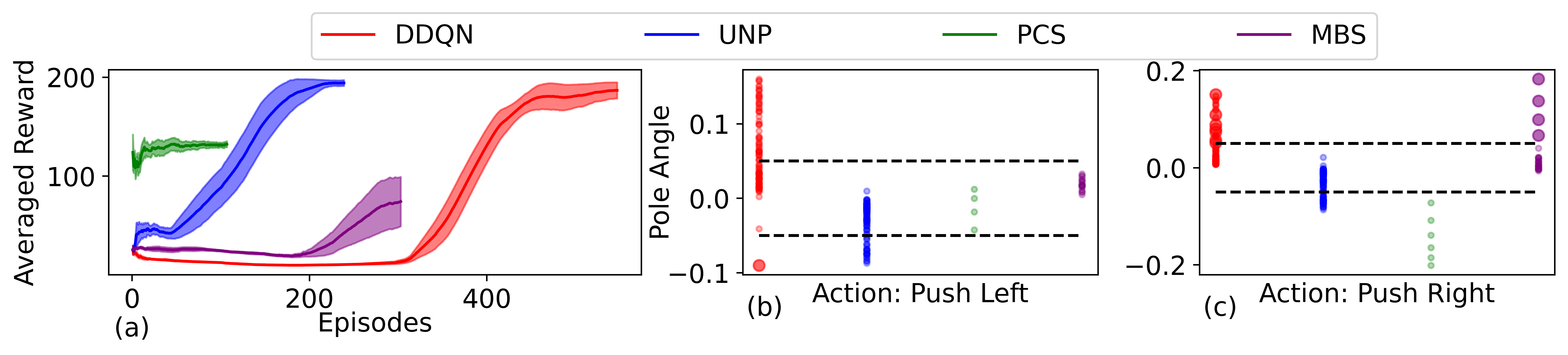}
  \vspace{-2mm}
  \caption{Experimental results for the CartPole environment. The left plot shows agent performance over time; the right plots each correspond to an action and show a test trace of the best performing policy for each agent. Transitions in which unsafe actions were taken have larger dots while all other transitions have small dots. The dashed lines indicate where the boundaries are between $S_{\text{danger}}$ and all other states. Note that the y-axis of the right plots is in radians.}
  \label{fig:cp_plot}
\end{figure*}

This section evaluates the proposed UNP shielding technique on three RL tasks, namely CartPole, Lane Keeping, and FlappyBird. In addition, we analyze their learning performances and compare them with the baselines DDPG and DDQN (see Section \ref{sec:bkgd}) and two recent shielding techniques as discussed below.

\textbf{PCS} This method, called Pre-Computation Shielding (PCS), generates a shield using pre-computation. It uses user-defined LTL properties and a two-player safety game to compute the set of safe states checked by the shield \citep{Alshiekh2018}.

\textbf{MBS} This method, called Model-Based Shielding (MBS), uses a given model and set of safe states to simulate trajectories into the future. If all simulated trajectories are found to result in unsafe states, then the action is rejected \citep{Giacobbe2021}.

We discuss random seed and hyper-parameter settings (see Tables \ref{tab:seed_info} and \ref{tab:hyperparams} in the Appendix), RL model architectures, and other experiment details (e.g. evaluation metrics, number of runs) in the Appendix.

\subsection{CartPole}

CartPole is a benchmark RL problem in which an agent must balance a pole attached to a cart. We used the environment from OpenAI Gym \citep{Brockman2016}. For CartPole, $x$ is the cart position, $\theta$ is the pole angle, and $\dot{\theta}$ is the pole angular velocity. Refer to Table \ref{tab:env_details} in the Appendix for more details. The reward function is the following:
\begin{equation}
    r =
  \begin{cases}
                                   1 & \text{while $|\theta| < 12^{\circ}$ and $|x| < 2.4$} \\
                                   0 & \text{otherwise} \\
  \end{cases}
\end{equation}
We define an unsafe behavior as moving the cart (i.e. the base of the pole) away from the direction of tipping only if the pole is already tipped past a certain degree. Consequently, this definition matches our conditions for bridging safety and optimality (i.e. Assumption \ref{assump}): exacerbating pole tipping through unsafe actions invariably will lead to problem failure and thus are also strictly sub-optimal. We define $A_{unp}(s)$ as
\begin{equation}
\begin{aligned}
    A_{unp,1}&(s) = \{\text{push cart left}\ |\ s \in S_{\text{danger},1}\} \\
    & \text{for } S_{\text{danger},1} = \{s \in S\ |\ \theta < -3^{\circ}\ \wedge\ \dot{\theta} < 0\} \\
    A_{unp,2}&(s) = \{\text{push cart right}\ |\ s \in S_{\text{danger},2}\} \\
    & \text{for } S_{\text{danger},2} = \{s \in S\ |\ \theta > 3^{\circ}\ \wedge\ \dot{\theta} > 0\}
\end{aligned}
\end{equation}
Note that $S_{\text{danger}}$ is different than $S_{\text{unsafe}}$ because it also
includes the set of states that can transition into $S_{\text{unsafe}}$.
We have chosen these thresholds of $|\theta| = 3^{\circ}$ and $\dot{\theta} = 0$ for $S_{\text{danger}}$ to be non-comprehensive and show that UNP shielding still results in safe optimal performance through the generalizability of RL. At action selection, the shield queries $A_{unp,1}(s)$ and $A_{unp,2}(s)$ and rejects the action if $a \in A_{unp}$ and $s \in S_{\text{danger}}$. The shield then uses the backup policy. For CartPole, the backup policy is simply the other available action.

\begin{figure*}[ht]
  \centering
  \includegraphics[width=0.95\textwidth]{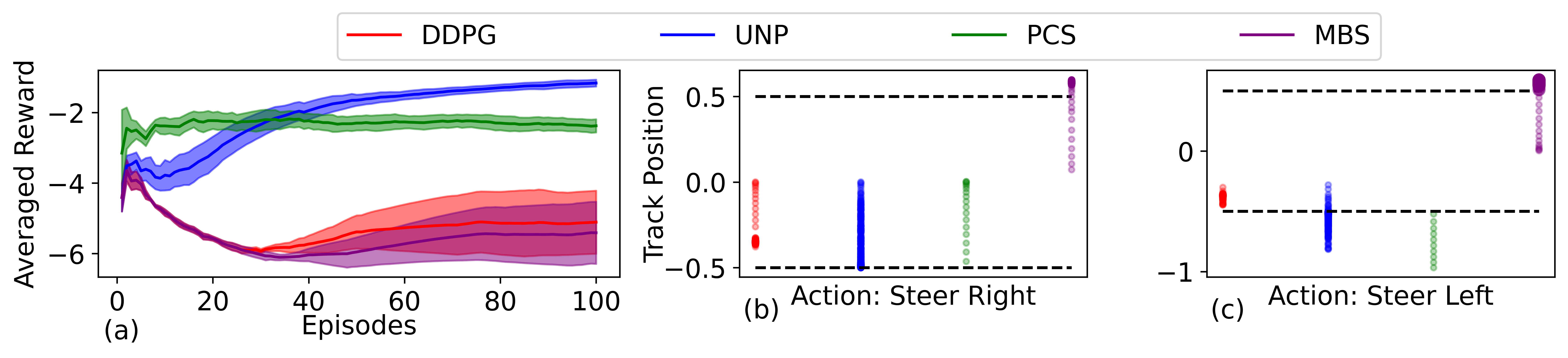}
    \vspace{-0.35cm}
  \caption{Experimental results for the Lane Keeping environment. For format details, see Figure \ref{fig:cp_plot}}
  \vspace{-0.35cm}
  \label{fig:torcs_plot}
\end{figure*}

\paragraph{Results} The results of the experiment are presented in Figure \ref{fig:cp_plot}.\footnote{Plotting stopped at maximum performance as many applications only want an optimal policy (which can be extracted and used) and do not care about destabilizing after further training.} Note that some actions may seem unsafe in Figure 1b because they are outside of the safe $\theta$ region but they are marked safe because $\dot{\theta}$ was not unsafe. Applying our UNP shield to CartPole allowed it to converge to the optimal solution in 200 episodes. DDQN still had not converged by 500 episodes.
In addition, the UNP plot depicts smaller error bars around the optimal behavior, implying not only quick but consistent convergence. From Figure 1b and 1c, UNP shielding successfully ensured safety.
DDQN failed to maintain safety, eventually using unsafe actions and leading to early episode termination.

Compared to PCS, our UNP method initially performed worse but learned significantly faster, eventually converging to superior behavior. Furthermore, UNP shielding did not require the definition of LTL properties, automata, or a security game. This saved significant implementation time. PCS performed poorly because our safety definition only involved the pole angle while the environment also considered cart position for failure.
Relying solely on the shield to learn satisfied safety but did not lead to good performance. PCS uses an update that assigns rewards to the original chosen action, not the shield-selected action. This leads to sub-optimal learned returns. Evidence of this can be seen in Figure 1c. PCS very quickly destabilized into a falling pole and failed to recover.
UNP completely outperformed MBS, both learning faster and converging to a better policy.
Because our safety specification was not comprehensive, MBS allowed actions deemed safe but actually led to failure. UNP is successful because it assigns rewards to actions already processed by the shield. These actions must be outside of $A_{unp}(s)$, guaranteeing not only safety but convergence to the optimal policy.

Readers may notice that, despite our following of Assumption \ref{assump}, results clearly did not show perfect behavior from when the agent began interacting with the environment. There are two reasons for this. First, elimination of unsafe or sub-optimal actions does not inherently solve the problem. The agent must still interact with the environment to learn the optimal policy and avoid actions with catastrophic consequences after a single step.
Second, shields can only guarantee safety for a single step. If the agent took an action that caused the system dynamics to inevitably transition the agent into $S_{\text{unsafe}}$ in the future, the shield would not be able to prevent this from happening (e.g. turning too late to stop the momentum of a vehicle).
UNP's results verify that our method accounts for both of these reasons, unlike previous state-based shielding methods, and leads to better safety and optimality.

\subsection{Lane Keeping}

The lane keeping task is a classic problem from the field of autonomous vehicles that requires an agent to keep a vehicle from driving off of the road \citep{Sallab2016}. Lane Keeping presents a greater challenge than CartPole because state information typically does not fully encapsulate the underlying physics involved in driving a vehicle. This ensures that simple policies based on one-step rewards are not sufficient for learning the optimal policy as momentum and track changes require multi-step behaviors. This version of the problem is simulated in TORCS \citep{EricEspie2005}. For Lane Keeping, $\delta$ is the vehicle position relative to the center of the road. Refer to Table \ref{tab:env_details} in the Appendix for more details. The reward function is the following:
\begin{equation}
  r =
  \begin{cases}
                                   0 & \text{while $-1 < \delta < 1$} \\
                                   -200 & \text{otherwise} \\
  \end{cases}
\end{equation}
Note that $\delta < 0$ corresponds to the right side of the track, $\delta > 0$ to the left side, $a < 0$ to right turns, and $a > 0$ to left turns. As actions leading the agent towards the edge of the road are both intuitively unsafe and cannot lead to optimal expected return, we can completely define $A_{unp}(s)$ using these actions. We formally define $A_{unp}(s)$ below
\begin{equation}
\begin{aligned}
    A_{unp,1}&(s)= \{a < 0\ |\ s \in S_{\text{danger},1}\} \\
    & \text{for } S_{\text{danger},1} = \{s \in S\ |\ \delta < -0.5\} \\
    A_{unp,2}&(s) = \{a > 0\ |\ s \in S_{\text{danger},2}\} \\
    & \text{for } S_{\text{danger},2} = \{s \in S\ |\ \delta > 0.5\}
\end{aligned}
\end{equation}
Similar to CartPole, the states in which $A_{unp}(s)$ is defined are non-comprehensive but we show that UNP shielding still results in safe optimal policies. The shield operates identically to that in CartPole except for the backup policy. For Lane Keeping, the backup policy uses the action value with the opposite sign. This corresponds to a turn in the opposite direction which must be safe as it steers the vehicle away from the edge.
\begin{equation}
    \pi_{shield}(s_t) = -\pi(s_t)
\end{equation}

\paragraph{Results} From Figure \ref{fig:torcs_plot}, we see that UNP resulted in a drastic performance increase. While DDPG struggled to reach a solution, the UNP agent converged in only 40 episodes. This improvement is further emphasized by the error bars in each plot. UNP shielding again maintained safe behavior throughout the test trace. Recall that UNP does not guarantee that agent's will not transition into unsafe states but does guarantee only safe actions are taken.
From Figure 2c, all actions in $S_{\text{danger}}$ are safe.
While DDPG did exhibit safe behavior in the test trace, it learned this policy only after significantly more episodes than the UNP shielded agent did.

\begin{figure*}[ht]
  \centering
  \includegraphics[width=0.95\textwidth]{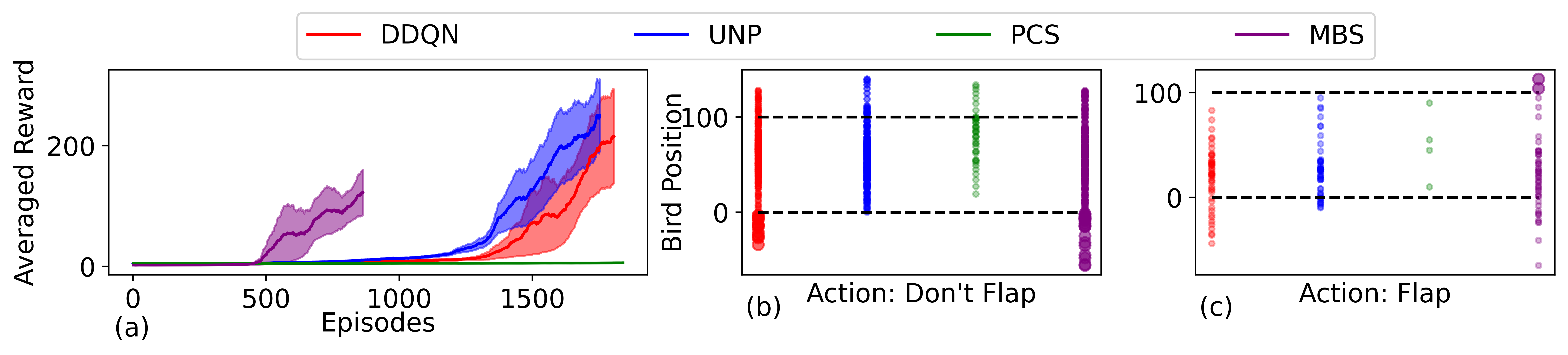}
  \caption{Experimental results for the FlappyBird environment. In the right plots, the black dashed lines mark the y-axis positions of the lower pipe (at $0$) and upper pipe (at $100$). For further format details, see Figure \ref{fig:cp_plot}}
  \vspace{-0.35cm}
  \label{fig:fp_plot}
\end{figure*}

For PCS, we had to implement a discrete form of Lane Keeping (five discrete actions instead of a continuous range).
PCS learned a competitive policy but, even in the simplified discrete environment, its maximum performance was still worse than UNP. From Figure 2b and 2c, PCS learned a safe policy.
However, the PCS agent tended to control the vehicle improperly, often allowing momentum to carry it very close to the right edge of the road.
We believe that this risky behavior resulted from the agent relying too much on the shield for control instead of learning the consequences of actions. This is a good example of how using a shield in some RL applications can be dangerous as the agent uses it as a crutch instead of as a method of learning safe optimal policies. UNP significantly outperformed MBS. The model-based rollouts used by MBS consistently found safe states in simulation even when the agent was dangerously close to the edge. This tricked the agent into believing the selected action was safe when it was actually unsafe and near failure. Figure 2b and 2c verify this behavior.
UNP was able to adaptively account for the shield's backup actions and learn the optimal policy within the episode limit.

\subsection{FlappyBird}

FlappyBird is a video game in which the agent must fly a bird through a series of pipes with gaps between them. After clearing a pair of pipes, the game generates a new set of pipes with a different gap location that the bird must adjust to. We used an implementation\footnote{https://github.com/yenchenlin/DeepLearningFlappyBird/} in PyGame (open-source). Note that we use state features in this problem. These features are detailed here:
\begin{align*}
    \phi = \{y_{\text{bird}}, x_{\text{dist\textunderscore to\textunderscore pipe}}, y_{\text{upipe}}, y_{\text{lpipe}}\}
\end{align*}
$y_{\text{bird}}$ is the bird's vertical location on the screen, $x_{\text{dist\textunderscore to\textunderscore pipe}}$ is the horizontal distance between the bird and the next pipe, $y_{\text{upipe}}$ is the top of the pipe gap's vertical location, and $y_{\text{lpipe}}$ is the bottom of the pipe gap's vertical location. Refer to Table \ref{tab:env_details} in the Appendix for more details. The reward function is the following:
\begin{equation}
  r =
  \begin{cases}
                                   1 & \text{if pipe cleared} \\
                                   0.1 & \text{else if bird is not contacting a pipe} \\
                                   -1 & \text{otherwise} \\
  \end{cases}
\end{equation}
The challenge in the FlappyBird environment is the large state space. Because we use pixels as states, the agent must learn how to generalize effectively across the state space to reach the optimal policy. As FlappyBird is a video game, unsafe transitions can be directly associated with failure. More specifically, we generalize unsafe and failure behaviors to any action above or below the pipe gap that lead the agent away from the pipe gap. We define $A_{unp}(s)$ below
\begin{equation}
\begin{aligned}
    A_{unp,1}&(s) = \{\text{no flap}\ |\ s \in S_{\text{danger},1}\} \\
    & \text{for } S_{\text{danger},1} = \{s \in S\ |\ y_{\text{bird}} < y_{\text{lpipe}}\}\\
    A_{unp,2}&(s) = \{\text{flap}\ |\ s \in S_{\text{danger},2}\} \\
    & \text{for } S_{\text{danger},2} = \{s \in S\ |\ y_{\text{bird}} > y_{\text{upipe}}\}\\
\end{aligned}
\end{equation}
In all other aspects, the shield operates identically to that in CartPole.

\paragraph{Results} Figure \ref{fig:fp_plot} shows the results of applying UNP as a shield in FlappyBird. As in CartPole and Lane Keeping, UNP successfully induced faster and more reliable convergence. UNP reached an average reward of 200, a score empirically verified to represent convergent behavior, in 1550 episodes while DDQN required 1750 episodes. Note that while this is a small difference in terms of episodes, it is a significant difference in wall-time (more than $20$ hours). Importantly, the agent was able to use state feature information with UNP to converge to the solution more quickly. Figure 3b and 3c show that UNP shielding successfully induced safe behavior.
Overall, the UNP agent learned the optimal policy significantly faster than the DDQN agent.

Both PCS and MBS performed sub-optimally compared to UNP. PCS failed to solve the problem. While it did succeed in preventing the agent from taking any unsafe actions, the shield in PCS again acted as a crutch for the agent. The PCS agent relied on the shield to keep it in safe states but it failed to learn how to act in them.
Combined with the aforementioned reward assignment procedure used in PCS, this led to the agent failing to learn the optimal policy. MBS learned significantly more than PCS and indeed learned faster than all of the other methods. However, it did not reach as high of performance as either UNP or DDQN did. The reason for this is the same as that in CartPole and Lane Keeping: the model used in MBS did not accurately signal that the agent was in unsafe situations. Figure 3b and 3c again support this conclusion.
UNP eventually outperformed both PCS and MBS while also preventing unsafe actions.

\section{Conclusion and Future Work}
\label{sec:conc}

We introduced safe permissibility and unsafe non-permissibility as a framework for connecting safety and optimality in RL and proposed a UNP shielding technique to enforce RL safety. Experimental evaluation on three RL tasks showed that UNP not only improves agent convergence speed to safe policies but also to optimal performance, and that it can deal with the challenges of safety and sample complexity in real-life RL problems through shielding and exploration limitation. Future work includes expanding permissibility to the environments mentioned in the Discussion (\ref{sec:method_disc} and investigating automatic shield construction, potentially via supervised learning of safe or unsafe states as in Type 1 permissibility.

\bibliographystyle{plainnat}
\bibliography{main}  






\newpage
\appendix

\centerline{\textbf{\large Appendix}}
\bigbreak

\section{Experiment and Analysis Details}

Results were collected using a set of five independent runs of each algorithm on each environment. The mean and standard deviation episode reward $R_{ep} = \sum_{i=0}^{T} r_i$ for timestep $i$ and episode ending time $T$ was taken of these five runs and used as the evaluation metric. This is a standard metric in RL representing the learning progress of an agent. The higher the episode reward over time (i.e. episodes), the faster the agent learned more optimal policies. Episode reward over episodes analysis allows for concise and informative conclusions on algorithm performance in RL. All experiments used a set of five random seeds corresponding with each run. Seeds were chosen randomly. The specific seeds for each environment are listed in Table \ref{tab:seed_info}.

\begin{table}[!htb]
  \caption{Random Seeds for Each Environment}
  \label{seeds-table}
  \centering
  \begin{tabular}{lc}
    CartPole & \{1234, 2000, 3000, 3456, 4500\} \\
    Lane Keeping & \{1234, 2000, 3000, 3456, 4500\} \\
    FlappyBird & \{1234, 1500, 2222, 3456, 5000\}
  \end{tabular}
  \label{tab:seed_info}
\end{table}

\section{Computational Setup}
\label{suppmat:comp_setup_desc}

All experiments were run on one of two setups. The first was a machine containing an NVIDIA GeForce GTX 1650 graphics card. The other was a machine containing an NVIDIA GeForce 840M graphics card. All experiments were run on the GPUs. Experimental implementations used the following tools: PyTorch 1.8.0+cu101, gym 0.15.4.

\section{Hyperparameters}
\label{suppmat:hparam_desc}

Hyperparameters were chosen based on either the original works of the specific environments applied or the original implementations (detailed in Table \ref{tab:hyperparams}). No search over values was performed. For reference, $\alpha$ is the learning rate, $t$ is the target network update rate, $\gamma$ is the discount factor, $\mathcal{N}$ is the number of exploration steps, $\epsilon$ is the exploration factor, $n$ is the batch size, and $\mathcal{H}$ is the replay buffer size.

\begin{table}[ht]
  \caption{Hyperparameters for Each Environment}
  \label{hparam-table}
  \centering
  \begin{tabular}{lccc}
    \toprule
     & CartPole & Lane Keeping & FlappyBird \\
    \midrule
    Critic $\alpha$ & 0.0005 & 0.0005 & 0.000005 \\
    Actor $\alpha$ & N/A & 0.0003 & N/A \\
    $t$ & 0.001 & 0.001 & 0.001 \\
    $\gamma$ & 0.99 & 0.99 & 0.99 \\
    $\mathcal{N}$ & 2000 & 2000 & 30000 \\
    $\epsilon$ & 0.995 & 0.999 & 0.99985 \\
    $n$ & 128 & 128 & 128 \\
    $\mathcal{H}$ & 50000 & 100000 & 50000 \\
    \bottomrule
  \end{tabular}
  \label{tab:hyperparams}
\end{table}

\section{Environment and Implementation Details}
\label{suppmat:env_dets}

Table \ref{tab:env_details} contains basic MDP information for our experimental environments. Further details are provided in the following subsections.

\begin{table*}[ht]
\small
\caption{Experiment Environment Descriptions}
\label{sample-table}
\centering
\begin{tabular}{lll}
\toprule
Environment & State Space & Action Space \\
\midrule
CartPole & 
 \begin{tabular}{lll}$x$: cart position; $\dot{x}$: cart velocity\\ $\theta$: pole angle; $\dot{\theta}$: pole angular velocity\\\end{tabular}
& $\{\text{push cart left}, \text{push cart right}\}$     \\ \hline
Lane Keeping     & 
\begin{tabular}{llll}$\theta$: vehicle angle; \\ $\delta$: vehicle position relative to center of road; \\ \{$v_{x}$, $v_{y}$, $v_{z}$\}: vehicle speed in x, y z dimensions\end{tabular}
& [-1, 1]: range of steering control \\ \hline
FlappyBird     & Pixels      & $\{\text{flap}, \text{don't flap}\}$  \\
\bottomrule
\end{tabular}
\label{tab:env_details}
\end{table*}

\subsection{CartPole}

The cart in CartPole is constrained to motion along the x-axis and the pole is attached at one end to the cart. An agent's goal is to keep the pole balanced vertically up and the cart between the boundaries for as long as possible. Failure termination occurs if the pole tips too far ($|\theta| \geq 12^{\circ}$) or the cart moves past boundaries on the x-axis ($|x| \geq 2.4$). Otherwise, episode length was 200 time steps. We used DDQN \citep{Mnih2015} for our implementation.

The Q-value function was approximated using a neural network. The network used fully connected layers of 16 and 32 nodes and output a Q-value for each possible action. ReLU activations were used for all internal layers. A target network was copied at the beginning of training and soft updated according to the target network update rate.

\subsection{Lane Keeping}

We used a simplified version of Lane Keeping in which the agent only needs to control steering. The agent's goal is to keep the vehicle on the road for as long as possible. Failure termination occurs if the vehicle moves off the edge of the road ($|\delta| > 1$). Otherwise, episode length was approximately 5000 time steps, equivalent to two laps on the simulated track. DDPG \citep{Lillicrap2016} was used as the base learning algorithm.

The critic and actor were modeled using neural networks. The critic network used fully connected layers of 128 and 256 nodes for the state input and a fully connected layer of 256 nodes for the action input. The outputs from these layers was concatenated and fed to another fully connected layer of 256 nodes. All layers used batch normalization and all internal layers used ReLU activations. The actor network used fully connected layers of 128 and 256 nodes. All layers used batch normalization. Internal layers used ReLU activations and the output layer used tanh activations. Target networks were copied at the beginning of training and soft updated according to the target network update rate.

\subsection{FlappyBird}

The agent's goal in FlappyBird is to navigate a bird through a series of gaps between pipes by flapping (or not flapping). FlappyBird is a survival game; an episode continues until the agent hits a pipe or it is manually cutoff. Failure termination occurs if the bird contacts a pipe. Note that the reward in this environment is directly connected to the agent's progress. We therefore set the episode cutoff to a reward of 1000 points, a score indicative of good performance and recognizable learning. DDQN was used as the base learning algorithm for this environment.

Due to the image-based state space, the Q-value function was approximated using a convolutional neural network. Inputs were stacks of the current frame and the previous three frames. Three convolutional layers were used to process this input: 32 filters of size 8 with stride 4 and max pooling with window size 2, 64 filters of size 4 with stride 2 and max pooling with window size 2, and 64 filters of size 3 with stride 1 and max pooling with window size 2. Padding was applied to ensure resultant image dimensions were powers of 2. The result was flattened and passed to two consecutive fully connected layers with 256 nodes each. ReLU actvations were used for the fully connected layers. A target network was copied at the beginning of training and soft updated according to the target network update rate.

\end{document}